\documentclass[11pt,letterpaper]{article}
\usepackage{emnlp2018}
\usepackage{times}
\usepackage{latexsym}

\usepackage[utf8]{inputenc}
\usepackage[T1]{fontenc}

\usepackage{amsmath,amssymb}
\usepackage{csquotes}
\usepackage{graphicx}
\usepackage{multirow}

\usepackage{CJKutf8}
\aclfinalcopy

\newcommand{\ja}[1]{\begin{CJK}{UTF8}{min}#1\end{CJK}}

\newcommand\eg{\textit{e.g.}\ }
\newcommand\ie{\textit{i.e.}\ }
\usepackage[acronym,shortcuts,acronymlists={hidden}]{glossaries}

\newacronym{nlp}{NLP}{Natural Language Processing}
\newacronym{nmt}{NMT}{Neural Machine Translation}
\newacronym{smt}{SMT}{Statistical Machine Translation}
\newacronym{mt}{MT}{Machine Translation}
\newacronym{lm}{LM}{Language Modeling}
\newacronym{bpe}{BPE}{Byte-Pair Encoding}
\newacronym{dataset}{MTNT}{Machine Translation of Noisy Text}
\newacronym{mtnt}{MTNT}{Machine Translation of Noisy Text}
\newacronym{enfr}{\texttt{en-fr}}{English-French}
\newacronym{fren}{\texttt{fr-en}}{French-English}
\newacronym{enja}{\texttt{en-ja}}{English-Japanese}
\newacronym{jaen}{\texttt{ja-en}}{Japanese-English}
\newacronym{oov}{OOV}{out-of-vocabulary words}

\newcommand{\emoji}[1]{\includegraphics[width=1em]{emoji_images/#1.png}}

\title{MTNT: A Testbed for Machine Translation of Noisy Text}

\author{Paul Michel \and Graham Neubig \\ 
  Language Technologies Institute\\ Carnegie Mellon University \\
  \texttt{\{pmichel1,gneubig\}@cs.cmu.edu}}

\date{}

\begin{document}

\maketitle

\begin{abstract}
Noisy or non-standard input text can cause disastrous mistranslations in most modern \ac{mt} systems, and there has been growing research interest in creating noise-robust \ac{mt} systems.
However, as of yet there are no publicly available parallel corpora of with naturally occurring noisy inputs and translations, and thus previous work has resorted to evaluating on synthetically created datasets.
In this paper, we propose a benchmark dataset for \ac{mtnt}, consisting of noisy comments on Reddit\footnote{\url{www.reddit.com}} and professionally sourced translations.
We commissioned translations of English comments into French and Japanese, as well as French and Japanese comments into English, on the order of 7k-37k sentences per language pair.
We qualitatively and quantitatively examine the types of noise included in this dataset, then demonstrate that existing MT models fail badly on a number of noise-related phenomena, even after performing adaptation on a small training set of in-domain data.
This indicates that this dataset can provide an attractive testbed for methods tailored to handling noisy text in MT.\footnote{The data is publicly available at \url{http://www.cs.cmu.edu/~pmichel1/mtnt/}.}
\end{abstract}

\section{Introduction}

\begin{displayquote}
\#nlproc is actualy f*ing hARD tbh \emoji{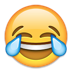}
\end{displayquote}

This handcrafted sentence showcases several types of noise that are commonly seen on social media: abbreviations (``\#nlproc''), typographical errors (``actualy''), obfuscated profanities (``f*ing''), inconsistent capitalization (``hARD''), Internet slang (``tbh'' for ``to be honest'') and emojis (\emoji{832}).  Although machine translation has achieved significant quality improvements over the past few years due to the advent of \ac{nmt} \cite{kalchbrenner-blunsom:2013:EMNLP,sutskever2014sequence,bahdanau2014neural,wu2016google}, systems are still not robust to noisy input like this \cite{belinkov2017synthetic,khayrallah2018noise}. For example, Google Translate\footnote{\url{translate.google.com} as of May 2018} translates the above example into French as:

\begin{displayquote}
\#nlproc est en train de f * ing dur hb
\end{displayquote}
which translates back into English as ``\#nlproc is in the process of [f * ing] hard hb''. This shows that noisy input can lead to erroneous translations that can be misinterpreted or even offensive.

Noise in social media text is a known issue that has been investigated in a variety of previous work \cite{eisenstein:2013:NAACL-HLT,baldwin-EtAl:2013:IJCNLP}. Most recently, \citet{belinkov2017synthetic} 
have focused on the difficulties that character based \ac{nmt} models have translating text with character level noise within individual words (from scrambling to simulated human errors such as typos or spelling/conjugation errors).
This is a good first step towards noise-robust \ac{nmt} systems, but as we demonstrate in \S\ref{sec:noise}, word-by-word replacement or scrambling of characters doesn't cover all the idiosyncrasies of language on the Internet.

At this point, despite the obvious utility of creating noise-robust MT systems, and the scientific challenges contained therein, there is currently a bottleneck in that there is no standard open benchmark for researchers and developers of \ac{mt} systems to test the robustness of their models to these and other phenomena found in noisy text on the Internet. 
In this work, we introduce \ac{mtnt}, a new, realistic dataset aimed at testing robustness of \ac{mt} systems to these phenomena.
The dataset contains naturally created noisy source sentences with professionally sourced translations both in a pair of typologically close languages (English and French) and distant languages (English and Japanese).
We collect noisy comments from the Reddit online discussion website (\S\ref{sec:collection}) in English, French and Japanese, and ask professional translators to translate to and from English, resulting in approximately 1000 test samples and from 6k to 36k training samples in four language pairs (\ac{enfr}, \ac{fren}, \ac{enja} and \ac{jaen}).
In addition, we release additional small monolingual corpora in those 3 languages to both provide data for semi-supervised adaptation approaches as well as noisy \ac{lm} experiments.
We test standard translation models (\S\ref{sec:mt_exp}) and language models (\S\ref{sec:lm_exp}) on our data to understand their failure cases and to provide baselines for future work.

\section{Noise and Input Variations in Language on the Internet}
\label{sec:noise}

\subsection{Examples from Social Media Text}

The term ``noise'' can encompass a variety of phenomena in natural language, with variations across languages (\eg what is a typo in logographic writing systems?) and type of content \cite{baldwin-EtAl:2013:IJCNLP}. To give the reader an idea of the challenges posed to \ac{mt} and \ac{nlp} systems operating on this kind of text, we provide a non-exhaustive list of types of noise and more generally input variations that deviate from standard \ac{mt} training data we've encountered in Reddit comments:  

\begin{itemize}
\item \textbf{Spelling/typographical errors}: ``across'' $\rightarrow$ ``accross'', ``receive'' $\rightarrow$ ``recieve'', ``could have'' $\rightarrow$ ``could of'', ``temps'' $\rightarrow$ ``tant'', ``\ja{除く}'' $\rightarrow$ ``\ja{覗く}''
\item \textbf{Word omission/insertion/repetition}: ``je n'aime pas'' $\rightarrow$ ``j'aime pas'',``je pense'' $\rightarrow$ ``moi je pense'' 
\item \textbf{Grammatical errors}: ``a ton of'' $\rightarrow$ ``a tons of'', 
``There are fewer people'' $\rightarrow$ ``There are less people''
\item \textbf{Spoken language}: ``want to'' $\rightarrow$ ``wanna'', ``I am'' $\rightarrow$ ``I'm'', ``je ne sais pas'' $\rightarrow$ ``chais pas'', ``\ja{何を笑っているの}'' $\rightarrow$ ``\ja{何わろてんねん}'', 
\item \textbf{Internet slang}: ``to be honest'' $\rightarrow$ ``tbh'', ``shaking my head'' $\rightarrow$ ``smh'', ``mort de rire'' $\rightarrow$ ``mdr'', ``\ja{笑}'' $\rightarrow$ ``w''/``\ja{草}''
\item \textbf{Proper nouns} (with or without correct capitalization): ``Reddit''$\rightarrow$ ``reddit''
\item \textbf{Dialects}: African American Vernacular English, Scottish, Provençal, Québécois, Kansai, Tohoku...
\item \textbf{Code switching}: ``This is so cute'' $\rightarrow$ ``This is so kawaii'', ``C'est trop conventionel'' $\rightarrow$ ``C'est trop mainstream'', ``\ja{現在捏造中…}'' $\rightarrow$ ``Now \ja{捏造}ing...''
\item \textbf{Jargon}: on Reddit: ``upvote'', ``downvote'', ``sub'', ``gild''
\item \textbf{Emojis and other unicode characters}: \emoji{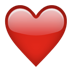},\emoji{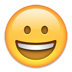},\emoji{832},\emoji{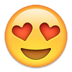},\emoji{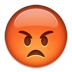}, \emoji{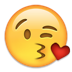}, \emoji{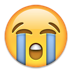}
\item \textbf{Profanities/slurs} (sometimes masked) ``f*ck'', ``m*rde'' \ldots
\end{itemize}

\subsection{Is Translating Noisy Text just another Adaptation Problem?}

To a certain extent, translating noisy text is a type of \emph{adaptation}, which has been studied extensively in the context of both \ac{smt} and \ac{nmt} \cite{axelrod-he-gao:2011:EMNLP,li-EtAl:2010:PAPERS3,luong2015stanford,chu-dabre-kurohashi:2017:Short,micelibarone-EtAl:2017:EMNLP2017,wang-EtAl:2017:EMNLP20174,michel2018extreme}. However, it presents many differences with previous domain adaptation problems, where the main goal is to adapt from a particular topic or style. In the case of noisy text, it will not only be the case that a particular word will be translated in a different way than it is in the general domain (e.g. as in the case of ``sub''), but also that there will be increased lexical variation (e.g. due to spelling or typographical errors), and also inconsistency in grammar (e.g. due to omissions of critical words or mis-usage).
The sum of these differences warrants that noisy \ac{mt} be treated as a separate instance than domain adaptation, and our experimental analysis in \ref{sec:mt:analysis} demonstrates that even after performing adaptation, \ac{mt} systems still make a large number of noise-related errors.

\section{Collection Procedure}
\label{sec:collection}

\begin{figure*}[!t]
\centering
\includegraphics[width=\textwidth]{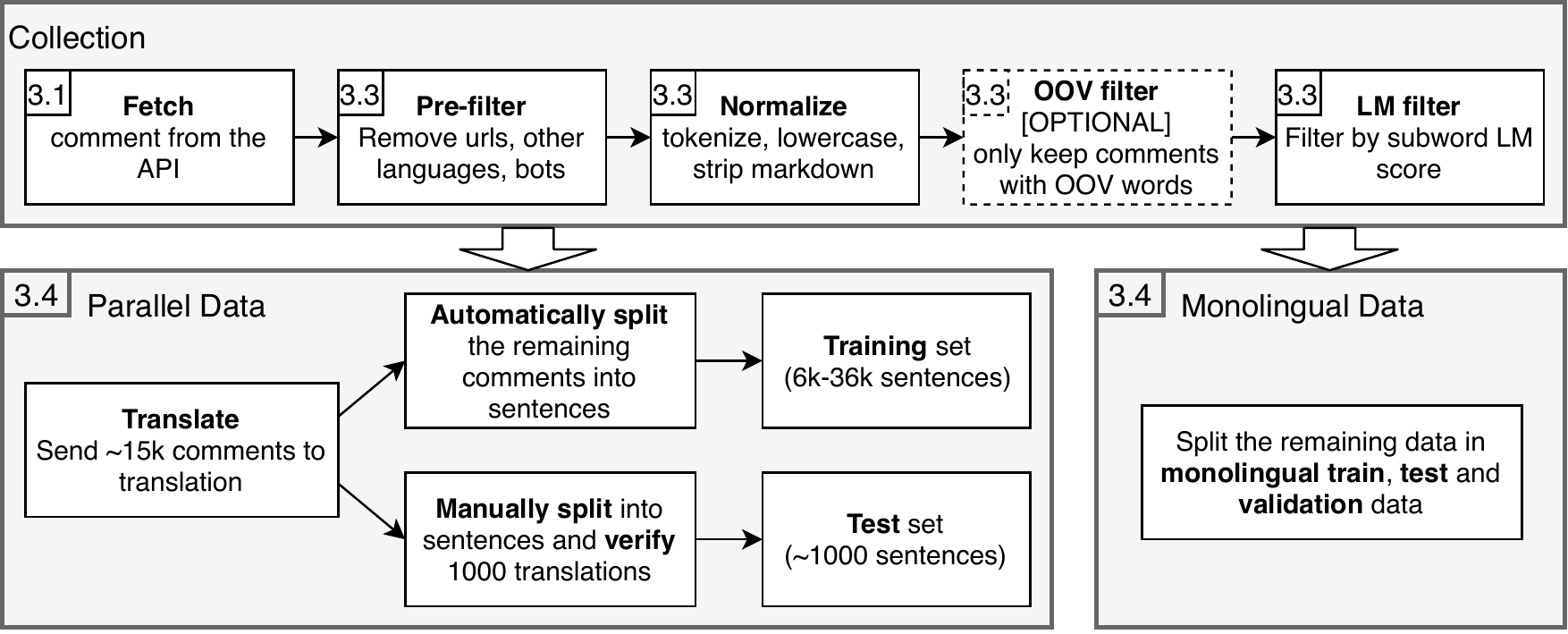}
\caption{\label{fig:mtnt_collection_diagram}Summary of our collection process and the respective sections addressing them. We apply the same procedure for each language.}
\end{figure*}

We first collect noisy sentences in our three languages of interest, English, French and Japanese. We refer to Figure \ref{fig:mtnt_collection_diagram} for an overview of the data collection and translation process.

We choose Reddit as a source of data because (1) its content is likely to exhibit noise, (2) some of its sub-communities are entirely run in different languages, in particular, English, French and Japanese, and  (3) Reddit is a popular source of data in curated and publicly distributed NLP datasets \cite{tan2016winning}. 
We collect data using the public Reddit API.
\footnote{In particular, we use this implementation: \url{praw.readthedocs.io/en/latest}, and our complete code is available at \url{http://www.cs.cmu.edu/~pmichel1/mtnt/}.} 

Note that the data collection and translation is performed at the comment level. We split the parallel data into sentences as a last step.

\subsection{Data Sources}

For each language, we select a set of communities (``subreddits'') that we know contain many comments in that language:

\begin{description}
\item[English:] Since an overwhelming majority of the discussions on Reddit are conducted in English, we don't restrict our collection to any community in particular.
\item[French:] \texttt{/r/france}, \texttt{/r/quebec} and \texttt{/r/rance}. The first two are among the biggest French speaking communities on Reddit. The third is a humor/sarcasm based offspring of \texttt{/r/france}.
\item[Japanese:] \texttt{/r/newsokur}, \texttt{/r/bakanewsjp}, \texttt{/r/newsokuvip}, \texttt{/r/lowlevelaware} and \texttt{/r/steamr}. Those are the biggest Japanese speaking communities, with over 2,000 subscribers.
\end{description}

We collect comments made during the 03/27/2018-03/29/3018 time period for English, 09/2018-03/2018 for French and 11/2017-03/2018 for Japanese. The large difference in collection time is due to the variance in comment throughput and relative amount of noise between the languages.

\subsection{Contrast Corpora}
\label{sec:ref_corp}

Not all comments found on Reddit exhibit noise as described in Section \ref{sec:noise}.
Because we would like to focus our data collection on noisy comments, we devise criteria that allow us to distinguish potentially noisy comments from clean ones.
Specifically, we compile a \emph{contrast} corpus composed of clean text that we can compare to, and find potentially noisy text that differs greatly from the contrast corpus. 
Given that our final goal is \ac{mt} robust to noise, we prefer that these contrast corpora consist of the same type of data that is often used to train \ac{nmt} models. We select different datasets for each language:

\begin{description}
\item[English:] The English side of the preprocessed parallel training data provided for the German-English WMT 2017 News translation task,\footnote{\url{http://www.statmt.org/wmt17/translation-task.html}} as provided on the website. This amounts to $\approx 5.85$ million sentences.
\item[French:] The entirety of the French side of the parallel training data provided for the English-French WMT 2015 translation task.\footnote{\url{http://www.statmt.org/wmt15/translation-task.html}}
This amounts to $\approx 40.86$ million sentences.
\item[Japanese:] We aggregate three small/medium sized \ac{mt} datasets: KFTT \cite{neubig11kftt}, JESC \cite{pryzant_jesc_2017} and TED talks \cite{cettoloEtAl:EAMT2012}, amounting to $\approx 4.19$ million sentences.
\end{description}

\subsection{Identifying Noisy Comments}

We now describe the procedure used to identify comments containing noise.

\paragraph{Pre-filtering} First, we perform three pre-processing to discard comments that do not represent natural noisy text in the language of interest:

\begin{enumerate}
\item Comments containing a URL, as detected by a regular expression.
\item Comments where the author's username contains ``bot'' or ``AutoModerator''. This mostly removes automated comments from bots.
\item Comments in another language: we run \texttt{langid.py}\footnote{\url{https://github.com/saffsd/langid.py}} \cite{lui-baldwin:2012:Demo} 
and discard comments where $p(\text{lang}\mid\text{comment}) > 0.5$ for any language other than the one we are interested in.%
\end{enumerate}

This removes cases that are less interesting, i.e. those that could be solved by rule-based pattern matching or are not natural text created by regular users in the target language. Our third criterion in particular discards comments that are blatantly in another language while still allowing comments that exhibit code-switching or that contain proper nouns or typos that might skew the language identification. In preliminary experiments, we noticed that these criteria 14.47, 6.53 and 7.09 \% of the collected comments satisfied the above criteria respectively.

\paragraph{Normalization}

After this first pass of filtering, we pre-process the comments before running them through our noise detection procedure.
We first strip Markdown\footnote{\url{https://daringfireball.net/projects/markdown}} syntax from the comments. For English and French, we normalize the punctuation, lowercase and tokenize the comments using the Moses tokenizer. For Japanese, we simply lowercase the alphabetical characters in the comments. Note that this normalization is done for the purpose of noise detection only. The collected comments are released without any kind of preprocessing.
We apply the same normalization procedure to the contrast corpora.

\paragraph{Unknown words} In the case of French and English, a clear indication of noise is the presence of \ac{oov}: we record all lowercased words encountered in our reference corpus described in Section \ref{sec:ref_corp} and only keep comments that contain at least one \ac{oov}.
Since we did not use word segmentation for the Japanese reference corpus, we found this method not to be very effective to select Japanese comments and therefore skipped this step.

\paragraph{Language model scores}
The final step of our noise detection procedure consists of selecting those comments with a low probability under a language model trained on the reference monolingual corpus. This approach mirrors the one used in \citet{moore-lewis:2010:Short} and \citet{axelrod-he-gao:2011:EMNLP} to select data similar to a specific domain using language model perplexity as a metric.
We search for comments that have a low probability under a sub-word language model for more flexibility in the face of \ac{oov} words. We segment the contrast corpora with \ac{bpe} using the sentencepiece\footnote{\url{https://github.com/google/sentencepiece}} implementation. We set the vocabulary sizes to $1,000$, $1,000$ and $4,000$ for English, French and Japanese respectively. We then use a 5-gram Kneser-Ney smoothed language model trained using \texttt{kenLM}\footnote{\url{https://kheafield.com/code/kenlm/}} \cite{Heafield-estimate} to calculate the log probability, normalized by the number of tokens for every sentence in the reference corpus. Given a reddit comment, we compute the normalized log probability of each of its lines under our subword language model. If for any line this score is below the 1st percentile of scores in the reference corpus, the comment is labeled as noisy and saved.

\subsection{Creating the Parallel Corpora}
\label{sec:parallel_corpora}

\begin{table}[tb]
\centering
\begin{tabular}{lccc}
& \#samples & \#src tokens & \#trg tokens \\ \hline
\texttt{en-fr} & 1,020 & 15,919 & 18,445\\
\texttt{fr-en} & 1,022 & 16,662 & 16,038\\
\texttt{en-ja} & 1,002 & 11,040 & 20,008\\
\texttt{ja-en} & 1,020 & 23,997 & 33,429\\
\hline
\end{tabular}
\caption{\label{tab:test_data_stats}Test set numbers.
}
\end{table}

Once enough data has been collected, we isolate $15,000$ comments in each language by the following procedure:
\begin{itemize}
\item Remove all duplicates. In particular, this handles comments that might have been scraped twice or automatic comments from bots.
\item To further weed out outliers (comments that are too noisy, \eg ASCII art, wrong language\ldots or not noisy enough), we discard comments that are on either end of the distribution of normalized \ac{lm} scores within the set of collected comments. We only keep comments whose normalized score is within the 5-70 percentile for English (resp. 5-60 for French and 10-70 for Japanese). These numbers are chosen by manually inspecting the data. 
\item Choose $15,000$ samples at random.
\end{itemize}

We then concatenate the title of the thread where the comment was found to the text and send everything to an external vendor for manual translations. Upon reception of the translations, we noticed a certain amount of variation in the quality of translations, likely because translating social media text, with all its nuances, is difficult even for humans. In order to ensure the highest quality in the translations, we manually filter the data to segment the comments into sentences and weed out poor translations for our test data. We thereby retain around $1,000$ sentence pairs in each direction for the final test set.

We gather the samples that weren't selected for the test sets to be used for training or fine-tuning models on noisy data. We automatically split comments into sentences with a regular expression detecting sentence delimiters, and then align the source and target sentences. Should this alignment fail (\ie the source comment contains a different number of sentences than the target comment after automatic splitting), we revert back to providing the whole comment without splitting. For the training data, we do not verify the correctness of translations as closely as for the test data. Finally, we isolate $\approx 900$ samples in each direction to serve as validation data. 

Information about the size of the data can be found in Table \ref{tab:test_data_stats},  \ref{tab:train_data_stats} and  \ref{tab:valid_data_stats} for the test, training and validation sets respectively. We tokenize the English and French data with the Moses \cite{Koehn:2007:MOS:1557769.1557821} tokenizer and the Japanese data with Kytea \cite{neubig11aclshort} before counting the number of tokens in each dataset.

\begin{table}[tb]
\centering
\begin{tabular}{lccc}
& \#samples & \#src tokens & \#trg tokens \\ \hline
\texttt{en-fr} & 36,058 & 841k & 965k \\
\texttt{fr-en} & 19,161 & 661k & 634k \\
\texttt{en-ja} & 5,775 & 281k & 506k\\
\texttt{ja-en} & 6,506 & 172k & 128k\\
\hline
\end{tabular}
\caption{\label{tab:train_data_stats}Training sets numbers. 
}
\end{table}

\begin{table}[tb]
\centering
\begin{tabular}{lccc}
& \#samples & \#src tokens & \#trg tokens \\ \hline
\texttt{en-fr} & 852 & 16,957 & 18,948 \\
\texttt{fr-en} & 886 & 41,578 & 46,886 \\
\texttt{en-ja} & 852 & 40,124 & 46,886\\
\texttt{ja-en} & 965 & 25,010 & 23,289\\
\hline
\end{tabular}
\caption{\label{tab:valid_data_stats}Validation sets numbers. 
}
\end{table}

\begin{table}[tb]
\centering
\begin{tabular}{clccc}
&&  \#samples & \#tok & \#char \\ \hline\hline
\multirow{2}{*}{\texttt{en}}& train&81,631&3,99M&18,9M\\
& dev&3,000& 146k&698k\\\hline
\multirow{2}{*}{\texttt{fr}}& train& 26,485 & 1,52M&7,49M\\
& dev&3,000&176k&867k\\\hline
\multirow{2}{*}{\texttt{ja}}& train&32,042&943k&3.9M\\
& dev&3,000&84k&351k\\
\hline
\end{tabular}
\caption{\label{tab:lm_data_stats}Monolingual data numbers.
}
\end{table}

\subsection{Monolingual Corpora}
\label{sec:lm_data}

After the creation of the parallel train and test sets, a large number of unused comments remain in each language, which we provide as monolingual corpora. This additional data has two purposes: first, it serves as a resource for in-domain training using semi-supervised methods relying on monolingual data (e.g. \citet{P16-1185,zhang-zong:2016:EMNLP2016}). Second, it provides a language modeling dataset for noisy text in three languages.

\begin{table*}[t]
\centering
\begin{tabular}{clcccc}
 & & Spelling & Grammar & Emojis & Profanities \\
\hline
\hline
\multirow{3}{*}{\texttt{en}} & newstest2014 & 0.210 & 0.189 & 0.000 & 0.030 \\
 & newsdiscusstest2015 & 0.621 & 0.410 & 0.021 & 0.076 \\
 & \ac{mtnt} (\ac{enfr}) & \bf 2.180 & \bf 0.559 & \bf 0.289 & \bf 0.239 \\
\hline
\multirow{3}{*}{\texttt{fr}} & newstest2014 & 2.776 & 0.091 & 0.000 & 0.245 \\
 & newsdiscusstest2015 & 1.686 & 0.457 & 0.024 & 0.354 \\
 & \ac{mtnt} & \bf 4.597 & \bf 1.464 & \bf 0.252 & \bf 0.690 \\
\hline
\multirow{4}{*}{\texttt{ja}}& TED & 0.011 & 0.266 & 0.000 & 0.000 \\
& KFTT & 0.021 & 0.228 & 0.000 & 0.000 \\
& JESC & 0.096 & 0.929 & 0.090 & \bf 0.058 \\
& \ac{mtnt} & \bf 0.269 & \bf 1.527 & \bf 0.156 & 0.036 \\
\hline
\end{tabular}
\caption{\label{tab:noise_qual} Numbers, per 100 tokens, of quantifiable noise occurrences. For each language and category, the dataset with the highest amount of noise is highlighted.}
\end{table*}

We select $3,000$ comments at random in each dataset to form a validation set to be used to tune hyper-parameters, and provide the rest as training data. The data is provided with one comment per line. Newlines within individual comments are replaced with spaces. Table \ref{tab:lm_data_stats} contains information on the size of the datasets. As with the parallel \ac{mt} data, we provide the number of tokens after tokenization with the Moses tokenizer for English and French and Kytea for Japanese.

\section{Dataset Analysis}
\label{sec:data_anal}

In this section, we investigate the proposed data to understand how different categories of noise are represented and to show that our test sets contain more noise overall than established \ac{mt} benchmarks.

\subsection{Quantifying Noisy Phenomena}

We run a series of tests to count the number of occurrences of some of the types of noise described in Section \ref{sec:noise}. Specifically we pass our data through spell checkers to count spelling and grammar errors. 
Due to some of these tests being impractical to run on a large scale, we limit our analysis to the test sets of \ac{mtnt}.

We use slightly different procedures depending on the tools available for each language. We test for spelling and grammar errors in English data using Grammarly\footnote{\url{https://www.grammarly.com/}}, an online resource for English spell-checking.
Due to the unavailability of an equivalent of Grammarly in French and Japanese, we test for spelling and grammar error using the integrated spell-checker in Microsoft Word 2013\footnote{\url{https://products.office.com/en-us/microsoft-word-2013}}. Note that Word seems to count proper nouns as spelling errors, giving higher numbers of spelling errors across the board in French as compared to English.

For all languages, we also count the number of profanities and emojis using custom-made lists and regular expressions\footnote{available with our code at \url{https://github.com/pmichel31415/mtnt}}. In order to compare results across datasets of different sizes, we report all counts per $100$ words.

The results are recorded in the last row of each section in Table \ref{tab:noise_qual}. In particular, for the languages with a segmental writing system, English and French, spelling errors are the dominant type of noise, followed by grammar error. Unsurprisingly, the former are much less present in Japanese.

\subsection{Comparison to Existing \ac{mt} Test Sets}

Table \ref{tab:noise_qual} also provide a comparison with the relevant side of established \ac{mt} test sets. For English and French, we compare our data to newstest2014\footnote{\label{notewmt15dev}\url{http://www.statmt.org/wmt15/dev-v2.tgz}} and newsdiscusstest2015\footnote{\label{notewmt15test}\url{http://www.statmt.org/wmt15/test.tgz}} test sets. For Japanese, we compare with the test sets of the datasets described in Section \ref{sec:ref_corp}.

Overall, \ac{mtnt} contains more noise in all metrics but one (there are more profanities in JESC, a Japanese subtitle corpus). This confirms that MTNT indeed provides a more appropriate benchmark for translation of noisy or non-standard text.

Compared to synthetically created noisy test sets \cite{belinkov2017synthetic} \ac{mtnt} contains less systematic spelling errors and more varied types of noise (\eg emojis and profanities) and is thereby more representative of naturally occurring noise.

\section{Machine Translation Experiments}
\label{sec:mt_exp}

We evaluate standard \ac{nmt} models on our proposed dataset to assess its difficulty. Our goal is not to train state-of-the art models but rather to test standard off-the-shelf \ac{nmt} systems on our data, and elucidate what features of the data make it difficult.

\subsection{Model Description}

All our models are implemented in DyNet \cite{neubig2017dynet} with the XNMT toolkit \cite{neubig2018xnmt}. We use approximately the same setting for all language pairs: the encoder is a bidirectional LSTM with 2 layers, the attention mechanism is a multi layered perceptron and the decoder is a 2 layered LSTM. The embedding dimension is 512, all other dimensions are 1024. We tie the target word embeddings and the output projection weights \cite{press-wolf:2017:EACLshort}. We train with Adam \cite{Kingma2014Adam} with XNMT's default hyper-parameters, as well as dropout (with probability $0.3$). We used \ac{bpe} subwords to handle \ac{oov} words. Full configuration details as well as code to reproduce the baselines is available at \url{https://github.com/pmichel31415/mtnt}.

\subsection{Training Data}
\label{sec:mt_train_data}

We train our models on standard \ac{mt} datasets:

\begin{itemize}
\item en $\leftrightarrow$ fr: Our training data consists in the europarl-v7\footnote{\url{http://www.statmt.org/europarl/}} and news-commentary-v10\footnote{\url{http://www.statmt.org/wmt15/training-parallel-nc-v10.tgz}} corpora, totaling $2,164,140$ samples, $54,611,105$ French tokens and $51,745,611$ English tokens (non-tokenized). We use the newsdiscussdev2015\textsuperscript{\ref{notewmt15dev}} dev set from WMT15 as validation data and evaluate the model on the newsdiscusstest2015\textsuperscript{\ref{notewmt15test}} and newstest2014\textsuperscript{\ref{notewmt15dev}} test sets.
\item en $\leftrightarrow$ ja: We concatenate the respective train, validation and test sets of the three corpora mentioned in \ref{sec:ref_corp}. In particular we detokenize the Japanese part of each dataset to make sure that any tokenization we perform will be uniform (in practice we remove ASCII spaces). This amounts to $3,900,772$ training samples ($34,989,346$ English tokens without tokenization). We concatenate the dev sets associated with these corpora to serve as validation data and evaluate on each respective test set separately.
\end{itemize}

\begin{table}[t]
\centering
\begin{tabular}{lcc}
& \ac{enfr}  & \ac{fren}   \\\hline\hline
newstest2014 & $33.52$ & $28.93$\\
newsdiscusstest2015 & $33.03$ & $30.76$\\ \hline
\ac{mtnt} & $21.77$ & $23.27$ \\
\ac{mtnt} (+tuning) & $29.73$ & $30.29$ \\
\hline
&\ac{enja}  & \ac{jaen} \\\hline\hline
TED&$14.51$&$13.25$\\
KFTT&$20.82$&$20.77$\\
JESC&$15.77$&$18.00$\\ \hline
\ac{mtnt}&$9.02$&$6.65$\\
\ac{mtnt} (+tuning) & $12.45$ & $9.82$ \\
\hline
\end{tabular}
\caption{\label{tab:bleu_scores} BLEU scores of \ac{nmt} models on the various datasets.}
\end{table}

\begin{table*}[t]
\centering
\begin{tabular}{ll}
\hline\hline
Source & Moi faire la gueule dans le métro me manque, c'est grave ? \\ \hline
Target & I miss sulking in the underground, is that bad? \\ \hline
Our model & I do not know what is going on in the metro, that is a serious matter. \\ \hline\
+ fine-tuning & I do not want to be in the metro, it's serious? \\ \hline\hline
Source & :o 'tain je me disais bien que je passais à côté d'un truc vu les upvotes. \\ \hline
Target &  :o damn I had the feeling that I was missing something considering the upvotes. \\ \hline
Our model & o, I was telling myself that I was passing over a nucleus in view of the Yupvoots. \\ \hline\
+ fine-tuning & o, I was telling myself that I was going next to a nucleus in view of the <unk>upvotes. \\ \hline\hline
Source & * C'est noël / pâques / pentecôte / toussaint : Pick One, je suis pas catho \\ \hline
Target &  Christmas / Easter / Pentecost / All Saints: Pick One, I'm not Catholic! \\ \hline
Our model & <unk> It is a pale/poward, a palec<unk>te d'<unk>tat: Pick One, I am not a catho! \\ \hline\
+ fine-tuning & <unk> It's no<unk>l / pesc<unk>e /pentecate /mainly: Pick One, I'm not catho! \\ \hline\hline
\end{tabular}
\caption{\label{tab:pre_post_finetunig_qual}Comparison of our model's output before and after fine-tuning in \ac{fren}.}
\end{table*}

\subsection{Results}
\label{sec:mt_results}

We use \texttt{sacreBLEU}\footnote{\url{https://github.com/mjpost/sacreBLEU}}, a standardized BLEU score evaluation script proposed by \citet{post2018call}, for BLEU evaluation of our benchmark dataset. It takes in detokenized references and hypotheses and performs its own tokenization before computing BLEU score. We specify the \texttt{intl} tokenization option. In the case of Japanese text, we run both hypothesis and reference through KyTea before computing BLEU score. We strongly encourage that evaluation be performed in the same manner in subsequent work, and will provide both scripts and an evaluation web site in order to facilitate reproducibility.

Table \ref{tab:bleu_scores} lists the BLEU scores for our models on the relevant test sets in the two language pairs, including the results on \ac{mtnt}.

\subsection{Analysis}
\label{sec:mt:analysis}

To better understand the types of errors made by our model, we count the n-grams that are over- and under- generated with respect to the reference translation. Specifically, we compare the count ratios of all 1- to 3-grams in the output and in the reference and look for the ones with the highest (over-generated) and lowest (under-generated) ratio.

We find that in English, the model under-generates the contracted form of the negative (``do not''/``don't'') or of auxiliaries (``That is''/``I'm''). Similarly, in French, our model over generates ``de votre'' (where ``votre'' is the formal 2nd person plural for ``your'') and ``n'ai pas'' which showcases the ``ne [\ldots] pas'' negation, often dropped in spoken language. Conversely, the informal second person ``tu'' is under-generated, as is the informal and spoken contraction of ``cela'', ``ça''. In Japanese, the model under-generates, among others, the informal personal pronoun \ja{俺} (``ore'') or the casual form \ja{だ} (``da'') of the verb \ja{です} (``desu'', to be). In \ac{jaen} the results are difficult to interpret as the model seems to produce incoherent outputs (\eg~``no, no, no\ldots'') when the \ac{nmt} system encounters sentences it has not seen before. The full list of n-grams with the top 5 and bottom 5 count ratios in each language pair is displayed in Table \ref{tab:compare_ngrams}.

\begin{table}[h]
\centering
\begin{tabular}{cccc}
\ac{fren} & \ac{enfr}&\ac{jaen} & \ac{enja} \\\hline\hline
\multicolumn{4}{c}{Over generated}\\\hline
\small{<unk>}&\small{<unk>} &\small{no, no,} & \small{\ja{※}}\\
\small{it is not}&\small{qu’ils} &\small{i} & \small{\ja{が }}\\
\small{I do not}&\small{de votre} &\small{no, no, no,} & \small{\ja{か ?}}\\
\small{That is}&\small{s’il} &\small{so on and} & \small{\ja{て }}\\
\small{not have}&\small{n’ai pas} &\small{on and so} & \small{\ja{す か ?}}\\
\hline
\multicolumn{4}{c}{Under generated}\\\hline
\small{it's} &\small{tu}& \small{|} & \small{\ja{？}}\\
\small{I'm} &\small{ça}& \small{Is} & \small{\ja{よ 。}}\\
\small{I don't} &\small{que tu} & \small{>} & \small{\ja{って}}\\
\small{>} &\small{!} & \small{""The} & \small{\ja{俺}}\\
\small{doesn't}&\small{as} & \small{those} & \small{\ja{だ 。}}\\
\hline
\end{tabular}
\caption{\label{tab:compare_ngrams}Over and under generated n-grams in our model's output for \ac{enfr}}
\end{table}

\subsection{Fine-Tuning}

Finally, we test a simple domain adaptation method by fine-tuning our models on the training data described in Section \ref{sec:parallel_corpora}. We perform one epoch of training with vanilla SGD with a learning rate of $0.1$ and a batch size of $32$. We do  not use the validation data at all. As evidenced by the results in the last row of Table \ref{tab:bleu_scores}, this drives BLEU score up by 3.17 to 7.96 points depending on the language pair. However large this increase might be, our model still breaks on very noisy sentences. Table \ref{tab:pre_post_finetunig_qual} shows three examples in \ac{fren}. Although our model somewhat improves after fine-tuning, the translations remain inadequate in all cases. In the third case, our model downright fails to produce a coherent output. This shows that despite improving BLEU score, naive domain adaptation by fine-tuning doesn't solve the problem of translating noisy text.

\section{Language Modeling Experiments}
\label{sec:lm_exp}

In addition to our \ac{mt} experiments, we report character-level language modeling results on the monolingual part of our dataset. We use the data described in Section \ref{sec:lm_data} as training and validation sets. We evaluate the trained model on the source side of our \ac{enfr}, \ac{fren} and \ac{jaen} test sets for English, French and Japanese respectively.

We report results for two models: a Kneser-Ney smoothed 6-gram model (implemented with KenLM) and an implementation of the AWD-LSTM proposed in \cite{merity2017regularizing}\footnote{\url{https://github.com/salesforce/awd-lstm-lm}}. We report the Bit-Per-Character (bpc) counts in table \ref{tab:bpc_lm}. We intend these results to serve as a baseline for future work in language modeling of noisy text in either of those three languages.

\begin{table}[tb]
\centering
\begin{tabular}{lcccc}
& \multicolumn{2}{c}{6-gram} &\multicolumn{2}{c}{AWD LSTM}\\
& dev &test & dev& test\\\hline\hline
English &2.081&2.179& 1.706& 1.810\\
French &1.906&2.090& 1.449&1.705\\
Japanese &5.003&5.497 & 4.801&5.225\\
\hline
\end{tabular}
\caption{\label{tab:bpc_lm}Language modeling scores}
\end{table}

\section{Related work}

Handling noisy text has received growing attention among various language processing tasks due to the abundance of user generated content on popular social media platforms \cite{crystal:01,herring:03,danet:07}. These contents are considered as noisy when compared to news corpora which have been the main data source for language tasks \cite{baldwin-EtAl:2013:IJCNLP,eisenstein:2013:NAACL-HLT}. They pose several unique challenges because 
they contain a larger variety of linguistic phenomena that are absent in the news domain and that lead to degraded quality when applying an model to out-of-domain data \cite{ritter2011named, luong2015stanford}.
Additionally, they are live examples of the Cmabrigde Uinervtisy (Cambridge University) effect, where state-of-the-art models become brittle while human's language processing capability is more robust \cite{sakaguchi2017robsut, belinkov2017synthetic}. 

Efforts to address these challenges have been focused on creating in-domain datasets and annotations \cite{owoputi2013improved, kong2014dependency, blodgett2017dataset}, and domain adaptation training \cite{luong2015stanford}. 
In \ac{mt}, improvements were obtained for \ac{smt} \cite{formiga-fonollosa:2012:POSTERS}. However, the specific challenges for neural machine translation have not been studied until recently \cite{belinkov2017synthetic, sperbertoward,cheng2018towards}. The first provides empirical evidence of non-trivial quality degradation when source sentences contain natural noise or synthetic noise within words, and the last two explore data augmentation and adversarial approaches of adding noise efficiently to training data to improve robustness. 

Our work also contributes to recent advances in evaluating neural machine translation quality with regard to specific linguistic phenomena, such as manually annotated test sentences for English to French translation, in order to identify errors due to specific linguistic divergences between the two languages \cite{isabelle-cherry-foster:2017:EMNLP2017}, or automatically generated test sets to evaluate typical errors in English to German translation \cite{sennrich:2017:EACLshort}. Our contribution distinguishes itself from this previous work and other similar initiatives \cite{peterson2011openmt12} by providing an open test set consisting of naturally occurring text exhibiting a wide range of phenomena related to noisy input text from contemporaneous social media.

\section{Conclusion}

We proposed a new dataset to test \ac{mt} models for robustness to the types of noise encountered in natural language on the Internet. We contribute parallel training and test data in both directions for two language pairs, English $\leftrightarrow$ French and English $\leftrightarrow$ Japanese, as well as monolingual data in those three languages. We show that this dataset contains more noise than existing \ac{mt} test sets and poses a challenge to models trained on standard \ac{mt} corpora. We further demonstrate that these challenges cannot be overcome by a simple domain adaptation approach alone. We intend this contribution to provide a standard benchmark for robustness to noise in \ac{mt} and foster research on models, dataset and evaluation metrics tailored for this specific problem.

\bibliography{references}
\bibliographystyle{acl_natbib_nourl}

\end{document}